\documentclass[review]{elsarticle}

\usepackage{lineno,hyperref}
\modulolinenumbers[5]

\usepackage{amsfonts}
\journal{Journal of \LaTeX\ Templates}

\begin{document}

\begin{frontmatter}

\title{Introducing user-prescribed constraints in Markov chains for nonlinear dimensionality reduction\tnoteref{mytitlenote}}

\author{Purushottam D. Dixit\fnref{myfootnote}}
\address{Department of Systems Biology, Columbia University, New York, NY}
\fntext[myfootnote]{email: dixitpd@gmail.com}

\begin{abstract}
Stochastic kernel based dimensionality reduction approaches have become popular in the last decade. The central component of many of these methods is a symmetric kernel that quantifies the vicinity between pairs of data points and a kernel-induced Markov chain on the data. Typically, the Markov chain is fully specified by the kernel through row normalization. However, in many cases, it is desirable to impose user-specified stationary-state and dynamical constraints on the Markov chain.  Unfortunately, no systematic framework exists to impose such user-defined constraints. Here, we introduce a path entropy maximization based approach to derive the transition probabilities of Markov chains using a kernel and additional user-specified constraints. We illustrate the usefulness of these Markov chains with examples.
\end{abstract}


\end{frontmatter}


\section{Introduction}

Recent technological advances allow collection of large amounts of high dimensional data across a range of scientific fields. Examples include gene expression levels in individual cells measured using single cell RNA sequencing~\citep{moon2017manifold}, pixel intensities in handwritten images~\citep{lecun1998mnist}, and collective neuronal firing data~\citep{sejnowski2014putting}.  It is quite often the case that the high dimensional data is generated from a small set of underlying factors. As a result, it is possible to embed the data in manifold in a much lower dimension.  A central task of dimensionality reduction methods is to identify these manifolds from sampled data points.

An important class of recently developed dimensionality reduction methods rely on a stochastic kernel based approach. Here, one starts with a positive and symmetric affinity kernel that reflects the proximity between pairs of data points. The kernel forms the basis of a Markov chain on the data. Finally, a lower dimensional representation is sought that preserves local neighborhoods of data points as quantified by the Markov chain. Popular examples of stochastic kernel based methods include diffusion maps~\citep{coifman2006diffusion} and Laplacian eigenmaps~\citep{belkin2003laplacian}, and  t-distributed stochastic neighbor embedding (t-SNE)~\citep{maaten2008visualizing}.

The Markov chain employed in these methods is obtained by row normalization of the kernel and is `local': the probability of transitioning from point `$a$' to another  point `$b$' depends only on the local neighborhood of $a$. Moreover, the kernel fully specifies both the stationary state distribution as well as the diffusion dynamics of the Markov chain. However, in many applications it is desirable to impose user-specified constraints on the Markov chain, for example, a prescribed stationary distribution. Here are some examples.

All atom molecular dynamics (MD) simulations are regularly used to explore the high dimensional conformational landscape of complex biomolecules. In many cases, the important dynamics can be projected on a lower dimensional `reaction coordinate'~\citep{hanggi1990reaction}. Diffusion maps have been used to identify these reaction coordinates from MD simulations~\citep{nadler2006diffusion,ferguson2010systematic,ferguson2011nonlinear}. In many cases, the ensemble of conformations obtained using an MD simulation may fail to reproduce experimental observables~\citep{pitera2012use,olsson2017combining,dixit2018caliber}. Thus, computational approaches are required to appropriately bias the Markov chain in order to obtain reaction coordinates that are consistent with experimental information.

Another example is from analysis of single cell gene expression data~\citep{moon2017manifold}. In any particular tissue of a multicellular organism,  cells are found in different stages of differentiation; ranging from stem cells to fully differentiated cells. In recent years, diffusion maps have been introduced to extract patterns of lineage differentiation from single cell gene expression profiles~\citep{haghverdi2015diffusion}. Recent work has stipulated that the gene expression profile of `stem-like' cells are more varied compared to the fully differentiated cells~\citep{teschendorff2017single} and assigning expression-dependent stationary probabilities to cells leads to a clearer identification of cell states and dynamics among those states~\citep{jin2018scepath}.

Notably, in these and other examples, there does not exist a systematic framework to manipulate the Markov chains used in stochastic kernel based dimensionality reduction methods according to user-prescribed constraints. Recently, we introduced a path-entropy based approach to modify Markov chains from constraints focusing on applications in biophysics and statistical physics~\citep{dixit2018perspective}. Here, one maximizes the entropy of the distribution over stochastic trajectories of the Markov chain subject to user-specified constraints.  We have used these path-entropy maximized Markov chains (PNMCs) to model statistical dynamics of biomolecular conformations~\citep{dixit2014inferring,dixit2015inferring}, to model biochemical reaction networks~\citep{dixit2018communication}, and to quantify inconsistencies in multicriteria decision making problems~\citep{dixit2018entropy}.

In this letter we first show that the transition probabilities associated with a row normalized Markov chain constitute a local maximum entropy probability distribution. In contrast, we seek a global maximum entropy Markov chain: we obtain the transition probabilities by maximizing the entropy of the ensemble of long stationary state paths of the Markov chain.   We illustrate the advantages of these path entropy maximized Markov chains using two examples: (1) constructing the phase transition associated reaction coordinate of a near-critical Ising model and (2) constructing the lineage differentiation tree of a cell population from single cell data.

\section{Background and notation\label{background}}

Below, we give a brief description of the traditionally used Markov chains in stochastic kernel based methods.

\subsection{Row normalized Markov chain (RNMC) on the data}

Consider $N$ data points $\{a, b, \dots \}$ in $\mathbb R^{{\rm n}}$.  A positive and symmetric kernel $\Delta(a,b) > 0$ is defined on all points $a$ and $b$ in $\mathbb R^{{\rm n}}$. A popular choice is the Gaussian kernel~\citep{coifman2006diffusion}
\begin{eqnarray}
\Delta(a,b) = \exp \left ( - \frac{d(a,b)^2}{2\varepsilon^2} \right ) \label{eq:DeltaDef}
\end{eqnarray}
where  $d(a,b)$ is the pairwise $\mathcal L_2$ distance. In Eq.~\ref{eq:DeltaDef}, $\varepsilon$ is the `bandwidth' of the kernel; $d(a,b) \gg \varepsilon \Rightarrow \Delta(a,b) \rightarrow 0.$  Alternative forms of the kernel have also been proposed. See for example~\citep{coifman2014diffusion,haghverdi2015diffusion,bermanis2016diffusion,moon2017visualizing}.

Typically, given a kernel, the Markov chain on data points is constructed as follows. First, we define $D(a) =\sum_b \Delta(a,b).$ $D(a)$ can be seen as a $\Delta-$kernel based density estimator at point $a$. Next,  an $\alpha-$dependent family of anisotropic kernels is introduced:
\begin{eqnarray}
\Delta^{(\alpha)}(a,b) = \frac{\Delta(a,b)}{D(a)^{\alpha} D(b)^{\alpha}} \label{eq:alphaD}
\end{eqnarray}
The $\alpha-$parametrized kernel is `row normalized' to obtain transition probabilities $q^{(\alpha)}_{ab}$
\begin{eqnarray}
q^{(\alpha)}_{ab} = \frac{\Delta^{(\alpha)}(a,b)}{Z^{(\alpha)}(a)} \label{eq:dmMarkov}
\end{eqnarray}
where $Z^{(\alpha)}(a) = \sum_b \Delta^{(\alpha)}(a,b)$ is the local partition function. The stationary distribution of the Markov chain is given by
\begin{eqnarray}
p^{(\alpha)}_a = \frac{Z^{(\alpha)}(a) }{\sum_a Z^{(\alpha)}(a) }. \label{eq:alphaStat}
\end{eqnarray}

The parameter $\alpha \in [0, 1]$  tunes the relative importance of the geometry of the lower dimensional manifold and the density statistics of data points on it~\citep{nadler2006diffusion}. For concreteness, consider that the data points $\bar x$ are generated according to a Fokker-Planck equation on some domain $\Omega \in \mathbb{R}^{n}$  with an equilibrium distribution $p(\bar x)~(\bar x\in \Omega)$. In the limit of infinitely many data points $N \rightarrow \infty$, the limiting diffusion process corresponding the backward operator of Eq.~\ref{eq:dmMarkov} approaches a Fokker-Planck equation as well~\citep{nadler2006diffusion}. Notably,  $\alpha$ controls its stationary distribution $\pi^{(\alpha)}(\bar x)$. Specifically, $\alpha = 0$ corresponds to a Fokker-Planck equation with a stationary distribution $\pi(\bar x) \propto p(\bar x)^2$ and $\alpha = 1/2$ corresponds to a Fokker-Planck equation with stationary distribution $\pi^{(\alpha)}(\bar x) = p(\bar x)$. In contrast, $\alpha = 1$ corresponds to a Fokker-Planck equation with a constant stationary distribution on $\Omega$~\citep{nadler2006diffusion}. We note that the correspondence only holds in the limit of infinite data. For example, the discrete time discrete state Markov chain defined by transition probabilities in Eq.~\ref{eq:dmMarkov} at $\alpha = 1$ does not have a uniform stationary distribution over the data points. For the rest of the manuscript, we omit the $\alpha-$dependence of the kernel for brevity and specify the value of $\alpha$ whenever necessary.

This row normalization procedure to obtain Markov chains from affinity kernels (or its closely related variants) is used in most stochastic kernel based methods including Diffusion maps~\citep{coifman2006diffusion}, Laplacian eigenmaps~\citep{belkin2003laplacian}, and tSNE~\citep{maaten2008visualizing}.

\subsection{RNMC is a local maximum entropy Markov chain}
The row normalized Markov chain described by the transition probabilities in Eq.~\ref{eq:dmMarkov} is in fact a local entropy maximized Markov chain. Consider that for each data point $a$ we want to find the transition probabilities $q_{ab}$ such that average squared distance traversed per unit time step is a specified number,  $\bar d^2(a)$. We maximize the entropy (conditioned on starting at data point $a$)~\citep{dixit2018perspective}
\begin{eqnarray}
\mathcal S_a = -\sum_b q_{ab} \log q_{ab}
\end{eqnarray}
subject to constraints\begin{eqnarray}
\sum_b q_{ab} = 1~{\rm and}~\sum_b q_{ab} d(a,b)^2 = \bar d^2(a). \label{eq:dist}
\end{eqnarray}
Entropy maximization subject to constraints in Eq.~\ref{eq:dist} yields the transition probabilities in Eq.~\ref{eq:dmMarkov} where $1/2\varepsilon^2$ is the Lagrange multiplier associated with the distance constraint.

\section{Path normalized markov chains (PNMC)}

How do we incorporate user-specified constraints in addition to the constraint in Eq.~\ref{eq:dist} on the Markov chain? For concreteness, let us denote by $\{ q_{ab} \}$ the transition probabilities of the sought Markov chain and $\{ p_a \}$ its stationary distribution. We may either {\it de novo} infer a Markov chain from specified constraints~\citep{dixit2014inferring,dixit2015inferring,dixit2015stationary} or obtain a least-deformed {\it updated} Markov chain with respect to a given {\it prior} Markov chain $\{ k_{ab} \}$ (with stationary distribution $\{ p_a\}$)~\citep{dixit2018caliber,dixit2018communication}.

A common approach to infer constrained stochastic processes is the dynamical version of the maximum entropy principle~\citep{dixit2018perspective}.  Consider a long stationary state paths $\Gamma \equiv  \cdots \rightarrow a_1 \rightarrow a_2 \rightarrow \cdots$ of duration $T \gg 1$ time steps of the Markov chain with hitherto unknown transition probabilities $\{ q_{ab} \}$. The first constraint we introduce (similar to Eq.~\ref{eq:dist}) is the path-ensemble average $\bar d^2$ of the squared distance traversed by a random walker on the data points. We have
\begin{eqnarray}
\bar d^2 &=& \frac{1}{T} \left ( \dots + d(a_{1},a_{2})^2 + d(a_{2},a_{3})^2 + \dots \right ) \nonumber \\ &\approx&  \sum_a p_a \sum_{b} q_{ab} d(a,b)^2 \label{eq:ave}
\end{eqnarray}
The second approximation holds in the limit $T\rightarrow \infty.$  In Eq.~\ref{eq:ave} $\{ p_a \}$ is the stationary distribution of the Markov chain. In addition other constraints of the form $\bar r = \sum_{a,b} p_a q_{ab} r_{ab}$ can also be introduced.

The maximum  entropy Markov chain or the path normalized Markov Chain (PNMC) is found as follows. The path entropy~\cite{dixit2014inferring,dixit2015inferring,dixit2015stationary,dixit2018perspective}
\begin{eqnarray}
\mathcal S = \sum_a p_a \mathcal S_a = -\sum_{a,b} p_a q_{ab} \log q_{ab}
\end{eqnarray}
is maximized subject to user-specified constraints using the method of Lagrange multipliers. We have the following constraints on the transition probabilities and the stationary distribution:
\begin{eqnarray}
\sum_b p_a q_{ab} &=& p_a, \sum_{a,b} p_a q_{ab} = 1, \sum_a p_a q_{ab}=  p_b \label{eq:c1}
\end{eqnarray}
and
\begin{eqnarray}
\sum_{a,b} p_a q_{ab} d(a,b)^2 = \langle d(a,b)^2 \rangle = \bar d^2. \label{eq:c2}
\end{eqnarray}
In addition, we also impose detailed balance
\begin{eqnarray}
p_aq_{ab} = p_b q_{ba}~\forall~a~{\rm and}~b.
\end{eqnarray}

At this stage, we have the option of constraining the stationary distribution~\citep{dixit2014inferring,dixit2015inferring}. Alternatively, we can maximize the entropy with respect to both the transition probabilities and the stationary distribution~\citep{dixit2015stationary}. Notably, these two choices lead to qualitatively different Markov chains.

\subsection{Unknown stationary distribution}

When the stationary distribution is not constrained, the  entropy is maximized with respect to both the transition probabilities as well as the stationary distribution. In this case, the transition probabilities are given by (see Appendix~\ref{ap_derivechain})~\citep{dixit2015stationary}
\begin{eqnarray}
q_{ab} = \frac{\nu_{1b}}{\eta_1 \nu_{1a}} \Delta(a,b) \label{eq:merw}
\end{eqnarray}
where $\eta_1$ is the Perron-Frobenius eigenvalue of $\Delta$ and $\bar \nu_1$ is the corresponding Perron-Frobenius eigenvector.  Note that since $\Delta$ is symmetric, the left and the right eigenvectors are identical. The stationary distribution resembles the ground state of the Schr{\"o}dinger's equation and is given by~\citep{dixit2015stationary}
\begin{eqnarray}
p_a  \propto\nu_{1a}^2 \label{eq:statdist}
\end{eqnarray}
We note that finding the Perron eigenvector of the kernel in constructing the PNMC in Eq.~\ref{eq:merw} does not add extra computational burden since estimation of the diffusion map also requires eigendecomposition of a matrix of the same size.

Notably, the PNMC with an unknown stationary distribution (Eq.~\ref{eq:merw}) can be recast as a RNMC corresponding to a modified symmetric and anisotropic kernel
\begin{eqnarray}
\Delta^{(\bar \nu_1)}(a,b) = \nu_{1a} \Delta(a,b) \nu_{1b} \label{eq:tnmp2}
\end{eqnarray}
From Eq.~\ref{eq:tnmp2} it is apparent that the PNMC defined by transition probabilities in Eq.~\ref{eq:merw} prefers to traverse in regions that are closely connected to each other as quantified by the eigenvector centrality~\citep{newman2008mathematics}.

\subsection{Imposing user-prescribed stationary distribution}

The maximum entropy Markov chain with a user-prescribed stationary distribution $\{ p_a \}$ and a constrained path-ensemble average $\bar d^2$ is given by (see Appendix~\ref{ap_derivechain})~\citep{dixit2014inferring,dixit2015inferring}
\begin{eqnarray}
q_{ab} = \frac{\rho_a \rho_b}{p_a} \Delta(a,b) \label{eq:pknown}
\end{eqnarray}
where $\Delta(a,b)$ is the same as Eq.~\ref{eq:DeltaDef}. The constants $\{ \rho_a \}$ are the fixed point of the nonlinear maxtrix equation
\begin{eqnarray}
R \Delta R \bar {\bf 1} = \bar p \label{eq:mefx}
\end{eqnarray}
where $R$ is the diagonal matrix with $R_{aa} = \rho_a$ and $\bar {\bf 1}$ is the column vector of ones. When the stationary probabilities are constrained to be equal, the transition probability matrix is symmetric and doubly stochastic. As above we denote by $1/2\varepsilon^2$ the Lagrange multiplier associated with $\bar d^2$. Interestingly, enforcing a uniform distribution on the Markov chain is sometimes termed as Sinkhorn normalization~\citep{idel2016review}. Indeed, it is known that converting a matrix into a doubly stochastic form may lead to better clustering performance~\citep{zass2007doubly}.  Moreover, various fast numerical algorithms have been proposed to solve Eq.~\ref{eq:mefx} with well established complexity bounds. See Idel~\citep{idel2016review} for a review.

We note that the path entropy based approach allows us to enforce a uniform stationary distribution $p_a = 1/N$ over data points even when $N$ is finite. We contrast this with the $\alpha-$dependent family of Markov chains with $\alpha = 1$ introduced above (see Eq.~\ref{eq:dmMarkov} and Eq.~\ref{eq:alphaStat}). The $\alpha = 1$ chain converges to a uniform distribution {\it only} in the limit $N\rightarrow \infty$.  Notably, the $N\rightarrow \infty$ limit of the PNMC with uniform distribution converges to the same Fokker-Planck equation as the $\alpha = 1$ limit~\citep{dixit2018caliber}.

\subsection{Updating a {\it prior} Markov chain}

Entropy maximization allows us to {\it update} a {\it prior} Markov chain. Consider that we have a prior Markov chain with transition probabilities $\{ k_{ab} \}$ and a stationary distribution $\{ p_a \}$ (see Dixit and Dill~\citep{dixit2018caliber} and Dixit~\citep{dixit2018communication} for more details). Instead of maximizing the path entropy, we minimize the Kullback-Leibler divergence~\citep{rached2004kullback}
\begin{eqnarray}
S = \sum_{a,b} p_a q_{ab} \log \frac{q_{ab}}{k_{ab}}
\end{eqnarray}
subject to the above constraints (Eq.~\ref{eq:c1} and Eq.~\ref{eq:c2}).

When the stationary distribution of the updated Markov chain is not constrained, its transition probabilities are given by~\citep{dixit2018communication}
\begin{eqnarray}
q_{ab} = \frac{\nu_{1b}}{\eta_1 \nu_{1a}} \Delta^*(a,b)
\end{eqnarray}
where
\begin{eqnarray}
\Delta^*(a,b) = \Delta(a,b) \sqrt{k_{ab}k_{ba}}, \label{eq:dprime}
\end{eqnarray}
 $\bar \nu_1$ is the Perron-Frobenius eigenvector of $\Delta^*$, and $\eta_1$ is the corresponding eigenvalue.

When the stationary distribution is constrained to a user-specified distribution $\{ p_a\}$, the transition probabilities of the Markov chain are given by
\begin{eqnarray}
q_{ab} = \frac{\rho_a\rho_b}{p_a}\Delta^*(a,b)
\end{eqnarray}
where $\Delta^*(a,b)$ is given by Eq.~\ref{eq:dprime} and $\bar \rho$ is the solution of the nonlinear equation
\begin{eqnarray}
R \Delta^* R \bar {\bf 1} = \bar p \label{eq:mefxprime}
\end{eqnarray}
where $R$ is the diagonal matrix with $R_{aa} = \rho_a$.

\subsection{Connection with optimal transport}

Finally, we discuss a curious connection with entropy-regularized optimal transport~\citep{peyre2017computational}. Optimal transport theory quantifies the `distance' between two distributions $\{ x_a \}$ and $ \{ y_a\}$ given a `cost matrix' $M$ as follows. First, one defines a set $U_{x,y}$ of positive matrices $P$ such that
\begin{eqnarray}
P \in U_{x,y} \Rightarrow \sum_a P_{ab} &=& y_b~\forall~b~{\rm and}~\sum_b P_{ab} = x_a~\forall~a \nonumber \\
\end{eqnarray}
The matrix $P_{ab}$ can be considered a joint probability matrix whose left and right marginals are $\{ x_a\}$ and $\{y_a\}$ respectively. The distance $\delta_M(x,y)$ is then given by
\begin{eqnarray}
\delta_M(x,y) &:=& \sum_{a,b} P_{ab}M_{ab}\nonumber \\
\rm{where}~P &=&{\rm argmin}~\sum_{a,b} P_{ab}M_{ab}. \label{eq:op}
\end{eqnarray}
Notably, while the problem in Eq.~\ref{eq:op} is a linear program, it can be regularized with an entropy function~\citep{cuturi2013sinkhorn}. Interestingly, the regularized problem is much faster to solve and can lead to better clustering of high dimensional data~\citep{cuturi2013sinkhorn}. The optimization problem in Eq.~\ref{eq:op} modifies to
\begin{eqnarray}
\delta_M^{\lambda}(x,y) &:=& \sum_{a,b} P^{\lambda}_{ab}M_{ab}\nonumber \\
\rm{where}~P^{\lambda} &=&{\rm argmin~} \sum_{a,b} P_{ab}M_{ab} - \lambda \sum_{a,b} P_{ab}\log P_{ab}. \nonumber \\ \label{eq:oph}
\end{eqnarray}

Note that if $x_a = y_a = p_a$, $P_{ab} = p_a q_{ab}$, and $M_{ab} = \Delta(a,b)$  then the problem in Eq.~\ref{eq:oph} is identical to the one of finding a Markov chain with a prescribed stationary distribution (see Eq.~\ref{eq:pknown}). In the future, it will be important to explore this connection further.

\section{Illustratations}

\subsection{Constructing the reaction coordinate of a near-critical Ising model}

\begin{center}
\begin{figure*}
        \includegraphics[scale=0.6]{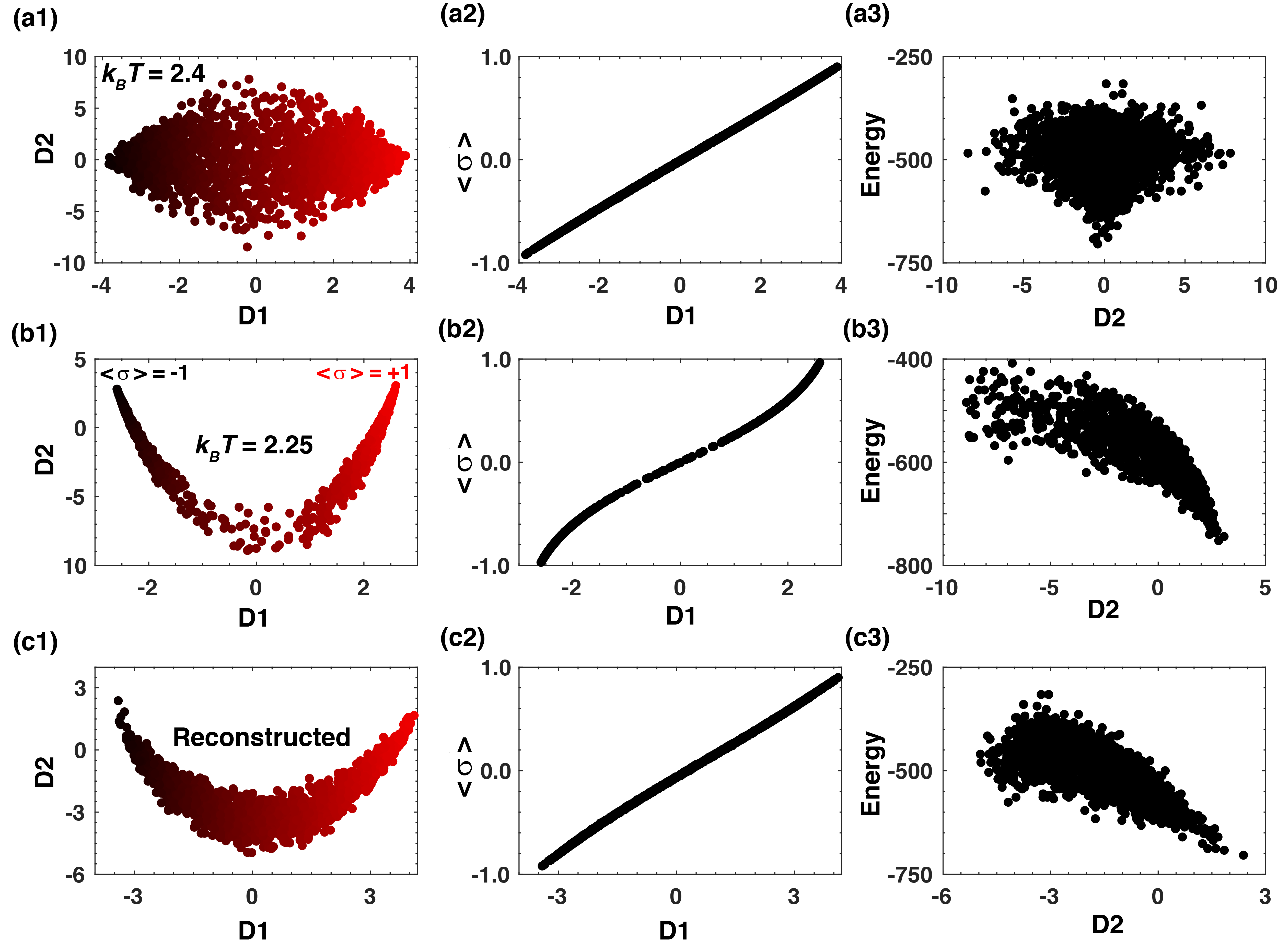}
        \caption{Panel (a1) The first two diffusion maps of a RNMC constructed using 2000 randomly sampled spin configurations of the Ising model at $k_BT = 2.4$. Panel (a2) The dependence of the average magnetization $\langle \sigma \rangle$ on the first diffusion map at $k_BT = 2.4$. Panel (a3) The dependence of the total energy on the second diffusion map at $k_BT = 2.4$. Panel (b1) The first two diffusion maps of a RNMC constructed using 2000 randomly sampled spin configurations of the Ising model at $k_BT = 2.25$. Panel (a2) The dependence of the average magnetization $\langle \sigma \rangle$ on the first diffusion map at $k_BT = 2.25$. Panel (a3) The dependence of the total energy on the second diffusion map at $k_BT = 2.25$. Panel (c1) The first two diffusion maps of a PNMC constructed using 2000 randomly sampled spin configurations of the Ising model at $k_BT = 2.4$ and then biased with a stationary distribution given in Eq.~\ref{eq:biasp}. Panel (c2) The dependence of the average magnetization $\langle \sigma \rangle$ on the first diffusion map of the PNMC. Panel (a3) The dependence of the total energy on the second diffusion map of the PNMC.\label{fg:ising}}
\end{figure*}
\end{center}

The two dimensional Ising model is a prototypical complex system studied extensively in statistical physics owing to its simplicity, analytical tractability, and rich phenomenology~\citep{chandler1987introduction}. Briefly, $N^2$ magnetic spins are arrange on an $N\times N$ square lattice. Individual spins $\sigma_i$ can take two values $\sigma_i = +1$ or $1.$ The probability of observing any state ${\bar \sigma} = \{ \sigma_1, \sigma_2, \dots , \sigma_{N^2} \}$ is given by the Gibbs-Boltzmann formula:
\begin{eqnarray}
p(\bar \sigma;\beta) = \frac{1}{Z(\beta)}\exp \left ( -\beta E(\bar \sigma) \right ) \label{eq:isingp}
\end{eqnarray}
where $\beta = 1/k_BT$ is the inverse temperature and the energy $E(\bar \sigma)$ is given by
\begin{eqnarray}
E(\bar \sigma) = \sum_{nn} \sigma_{i}\sigma_j. \label{eq:isinge}
\end{eqnarray}
The summation in Eq.~\ref{eq:isinge} is taken over all pairs of nearest neighbors on the lattice. In the limit $N\rightarrow \infty$, the Ising model exhibits a phase transition at $T \approx 2.26$ below which the average magnetization $\langle \sigma \rangle$ either assume the value $\langle \sigma \rangle \approx +1$ or $\langle \sigma \rangle \approx -1$ with an infinitely high barrier to flip the sign of all the spins~\citep{chandler1987introduction}.  Notably, the energy barrier to flip all spins is finite when $N$ is finite.

To understand whether the trajectories in the spin configuration space that lead to a complete reversal of average magnetization can be collapsed onto a reaction coordinate  we perform two calculations with the Ising model, one away from the critical points at $k_BT = 2.4$ and one near the critical point at $k_BT = 2.25$ respectively. The number of spins in the Ising model was $20\times 20 = 400$.  At each temperature, we sample 2000 spin configurations using Eq.~\ref{eq:isingp}. Next, we constructed a Gaussian kernel where the distance $d(a,b)$ between two spin configurations $\bar \sigma_a$ and $\bar \sigma_b$ was defined as the $\mathcal L_2$ distance between the two configurations.  We chose $\alpha = 0$ (see Eq.~\ref{eq:dmMarkov}) and $\varepsilon$ chosen to bethe 10$^{\rm th}$ percentile of all pairwise distances between spin configurations. Next we constructed the RNMC according to Eq.~\ref{eq:dmMarkov} and obtained its first two diffusion maps. Panel (a1) and panel (b1) in Fig.~\ref{fg:ising} show these  two diffusion maps for the two temperatures. Colors of individual points shows the net magnetization per spin ($\langle \sigma\rangle = -1$ in black and $\langle \sigma\rangle = 1$ in red). It is clear that the first diffusion map $D_1$ correlates strongly with the average magnetization $\langle \sigma \rangle$ (panels (a2) and (b2) of Fig.~\ref{fg:ising}). Interestingly, the second diffusion map correlates with the total energy close to the critical point ($k_BT=2.25$) but not away from the critical point ($k_BT = 2.4$)  (panels (a3) and (b3) of Fig.~\ref{fg:ising}). In other words, the diffusion map approach identifies the average spin $\langle \sigma\rangle $ and the total energy $E(\bar \sigma)$ as the two important `reaction coordinates/ of the Ising model close to the critical temperature.

Next, we examine whether we can reconstruct the reaction coordinate near the critical temperature using the simulation at $k_BT = 2.4$. The spin configurations  sampled at $k_B T = 2.4$ are distributed according to Eq.~\ref{eq:isingp} with $\beta = 1/2.4$. In order to reproduce the near critical behavior at $k_BT = 2.25$, we constructed a PNMC as follows. As above, we first constructed a Gaussian kernel with $\alpha = 0$ and $\varepsilon$ chosen to be the $10^{\rm th}$ percentile of all pairwise distances. Next, we imposed a stationary distribution on the PNMC given by
\begin{eqnarray}
p(\bar \sigma) \propto \exp \left ( -\left ( \frac{1}{2.25} - \frac{1}{2.4} \right ) E(\bar \sigma) \right ). \label{eq:biasp}
\end{eqnarray}
$p(\bar \sigma)$ in Eq.~\ref{eq:biasp} corresponds to the additional weight each data point must receive in order to reflect the near critical behavior. Notably, as seen in panel (c1) of Fig.~\ref{fg:ising}, the biasing captures the shape of the two dimensional reaction coordinate. Moreover, Panels (c2) and (c3) show that in agreement with the simulation near the critical temperature (panesl (b1), (b2), and (b3)), the first diffusion map of the PNMC corresponds to the average magnetization and the second diffusion map corresponds to the total energy. In other words, the biased PNMC is able to reproduce the reaction coordinates close to the critical temperature.

As noted in the introduction, diffusion maps are often used to explore the lower dimensional reaction coordinates from conformations sampled from MD simulations~\citep{ferguson2011nonlinear}. Our work suggests that imposing user-prescribed constraints on the Markov chain  allows us to systematically explore the dependency of the reaction coordinates on user-specified biases.

\subsection{Single cell gene expression in mouse haematopoietic stem cells~\citep{moignard2013characterization}}

\begin{figure*}
        \includegraphics[scale=0.65]{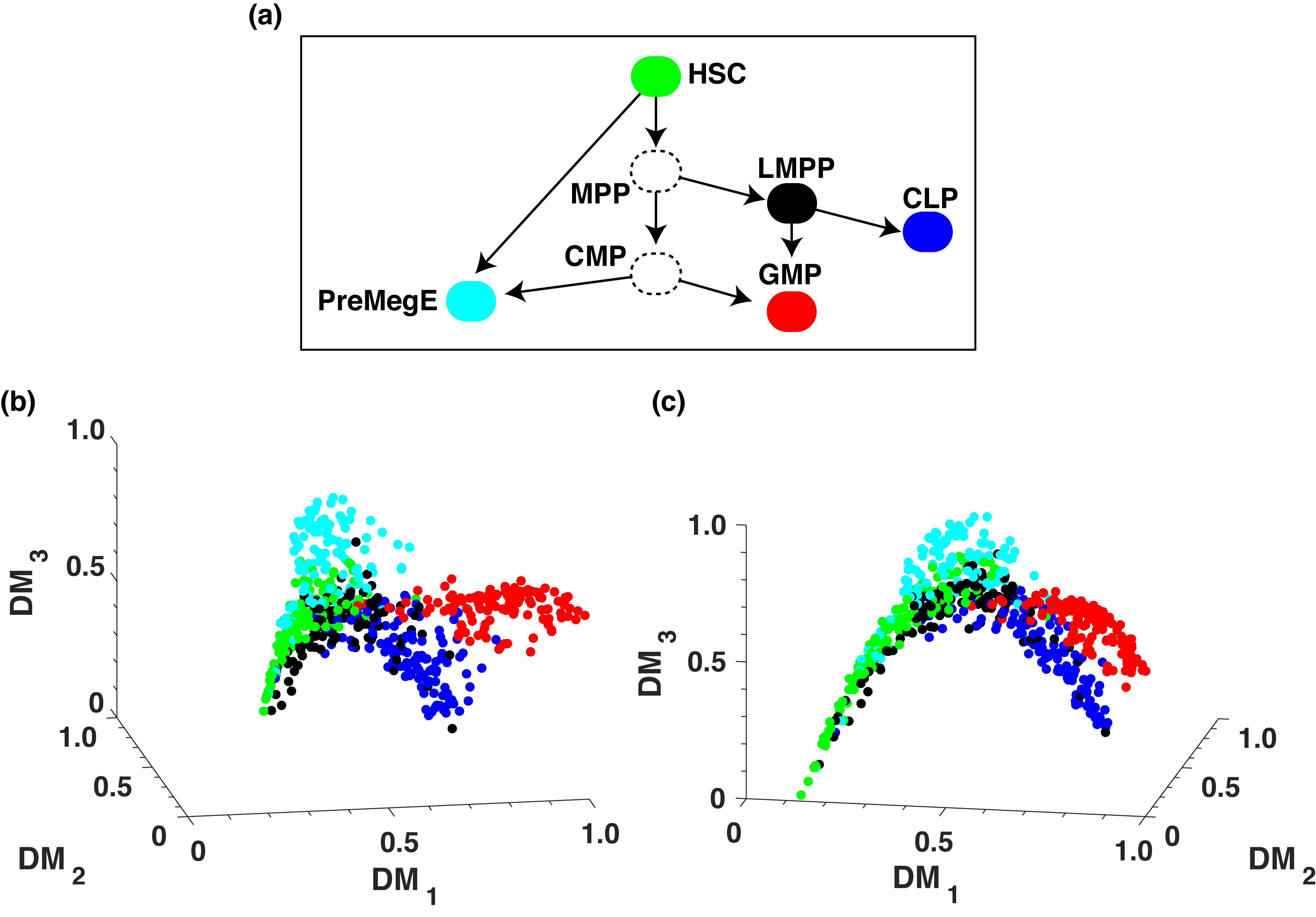}
        \caption{Panel (a) A biologically established differentiation trajectory of HSCs. Panel (b) The differentiation trajectory elucidated using the PNMC-derived diffusion maps. Panel (c) The differentiation trajectory elucidated using the RNMC-derived diffusion maps. \label{fg:diff}}
\end{figure*}

Next, we looked at cell transcription factor abundance profiles at the single cell level in mouse haematopoietic stem cells~\citep{moignard2013characterization}. Data was collected on 597 cells from five different cell types: haematopoietic stem cell (HSC), lymphoid-primed multipotent progenitor (LMPP), megakaryocyte-erythroid progenitor (PreMegE), common lymphoid progenitor (CLP) and granulocyte–monocyte progenitor (GMP). The known differentiation map~\citep{moignard2013characterization} of the cell types is given in Fig.~\ref{fg:diff}.

We constructed a symmetric kernel on the data points using the recently introduced  PHATE ({\bf P}otential of {\bf H}eat-diffusion for {\bf A}ffinity-based {\bf T}ransition {\bf E}mbedding) approach~\citep{moon2017visualizing}. We used $k=5$ nearest neighbors and the shape parameter equal to $\beta = 8$. The kernel is given by
\begin{eqnarray}
\Delta^{({k,\beta})}(a,b) = \exp \left ( -\left (\frac{d(a,b)}{\varepsilon_k(a) } \right )^{\beta}\right) + \exp \left ( -\left (\frac{d(a,b)}{\varepsilon_k(b) } \right )^{\beta}\right) \nonumber \\ \label{eq:moon}
\end{eqnarray}
where $\varepsilon_k(a)$ is the distance of the $k^{\rm th}$ nearest neighbor from data point $a$.

We constructed an RNMC using the phate kernel with $\alpha = 0$ in Eq.~\ref{eq:dmMarkov}. We also constructed a PNMC with a biased stationary distribution. It is argued that the abundance  profiles of individual cells are more variable in the stem cell-state compared to the fully differentiated state~\citep{teschendorff2017single,jin2018scepath}. Moreover, it is stipulated that assigning an abundance-variability dependent stationary distribution to cells may lead to a better resolution of cell state dynamics~\citep{jin2018scepath}. Accordingly, we imposed a stationary distribution on the PNMC that related to the entropy of the abundance profile of the 18 transcription factors. If $x_{ij}$ was the abundance of factor $j$ in cell $i$, we estimated the entropy
\begin{eqnarray}
s_i = -\sum_j x_{ij} \log x_{ij}.
\end{eqnarray}
The stationary distribution for a cell $i$ was set to
\begin{eqnarray}
p_i \propto \frac{1}{1 + \exp(-s_i)}.
\end{eqnarray}
This stationary distribution favors cells with higher entropy.

In Fig.~\ref{fg:diff}, we show the diffusion maps constructed using the PNMC (panel (a) of Fig.~\ref{fg:diff}) and the RNMC (panel (b) of Fig.~\ref{fg:diff}). Notably, individual branches of the differentiation profile are better resolved with the PNMC. For example, the PreMegE cell type is better separated from its the stem cell HSC (green$\rightarrow$ cyan). In the same vain, the average distance between different cell type clusters was higher for the PNMC by $\sim 10\%$ compared to the RNMC (paired t-test $p \sim 2\times 10^{-7}$).

\section{Concluding discussion}

In typical stochastic kernel based approaches, a `local' Markov chain (RNMC) is constructed on the data points via  row normalization of a positive and symmetric kernel. In this article we introduced a global path-entropy maximization based alternative normalization approach. Notably, both the stationary distribution and the diffusive properties of the path-entropy normalized Markov chain (PNMC) can be explicitly controlled by the user allowing a much greater flexibility with respect to the long-time properties of the Markov chain compared to the typical row normalized Markov chain. We showed how imposition of user-prescribed constraints on the PNMC can be leveraged to gain more insights about the data, for example, in predicting reaction coordinates in complex systems or in deciphering the differentiation trajectory of cells from single cell data.

On the one hand, the Markov chains introduced here maximize the path entropy over very long stationary state trajectories. This may induce attraction between distant data points that have a high connectivity (see Eq.~\ref{eq:tnmp2}). On the other hand, the row normalized Markov chain typically used in diffusion maps represents a maximum entropy Markov chain over a single time step. A straightforward generalization of the current work is to consider entropy maximization over a finite number of steps~\citep{frank2014information}. Notably, recent work suggests that incorporating finite path statistics may improve the quality of dimensionality reduction. For example, Little et al.~\citep{little2017path} have shown that modifying the definition of the pairwise distance to include the connectivity between data points can lead to better embedding properties specifically for clusters of variable shapes.  Steinerberger~\citep{steinerberger2016filtering} showed that optimally choosing transition probabilities from Markov chains of multiple path-sizes effectively filters out unconnected data points.

{\bf Acknowledgment}: I would like to thank Dr. Manas Rachh and Dr. Stefan Steinerberger for fruitful discussions about the manuscript. I would also like to thank Prof. Ronald Coifman for pointing out the analogy with optimal transport.

\section*{References}


\newpage
\pagebreak

\setcounter{figure}{0}
\makeatletter
\renewcommand{\thefigure}{A\@arabic\c@figure}
\makeatother

\setcounter{equation}{0}
\makeatletter
\renewcommand{\theequation}{A\@arabic\c@equation}
\makeatother

\setcounter{section}{0}
\makeatletter
\renewcommand{\thesection}{A\@arabic\c@section}
\makeatother

\section{Derivation of transition probabilities\label{ap_derivechain}}

\subsection{When the stationary probabilities are constrained}

We maximize the trajectory entropy
\begin{eqnarray}
\mathcal S = \sum_a p_a \mathcal S_a = -\sum_{a,b} p_a q_{ab} \log q_{ab}
\end{eqnarray}
subject to constraints
\begin{eqnarray}
\sum_b p_a q_{ab} &=& p_a, \sum_{a,b} p_a q_{ab} = 1, \sum_a p_a q_{ab}=  p_b \label{eq:ac1}
\end{eqnarray}
and
\begin{eqnarray}
\sum_{a,b} p_a q_{ab} d(a,b)^2 = \langle d(a,b)^2 \rangle = \bar d^2. \label{eq:ac2}
\end{eqnarray}
We solve the constrained optimization problem using the method of Lagrange multipliers. We write the unconstrained optimization function
\begin{eqnarray}
\mathcal C &=& S + \sum_a l_a \left ( \sum_b p_aq_{ab} - p_a \right) + \sum_b m_b \left ( \sum_a p_aq_{ab} - p_b \right) \nonumber \\
&-& \frac{1}{2\varepsilon^2}\left (\sum_{a,b} p_a q_{ab} d(a,b)^2 -  \bar d^2  \right )
\end{eqnarray}

\begin{eqnarray}
0 &=& -(\log q_{ab} + 1) + l_a +  m_b -  \frac{d^2(a,b)}{2\varepsilon^2}\\
\Rightarrow q_{ab} &=& \frac{\rho_a \lambda_b}{p_a}\exp \left( -\frac{d(a,b)^2}{2\varepsilon^2}\right )
\end{eqnarray}
where $e^{l_a - 1} = \rho_a/p_a$ and $e^{m_b} = \lambda_b$. Notably, since $p_a q_{ab} = p_b q_{ba}$, we also have $\rho_a = \lambda_a~\forall~a$. Thus, the transition probabilities are given by~\citep{dixit2014inferring}
\begin{eqnarray}
q_{ab} = \frac{\rho_a\rho_b}{p_a} exp \left( -\frac{d(a,b)^2}{2\varepsilon^2}\right )
\end{eqnarray}

\subsection{When the stationary probabilities are not constrained}

When the stationary probabiities are not constrained, we maximize the unconstrained optimization function with respect to $q_{ab}$ as well as $p_a$. Moreover, we have an additional constraint
\begin{eqnarray}
\sum_{a,b} p_a q_{ab} = 1.
\end{eqnarray}

We write the unconstrained optimization function as above
\begin{eqnarray}
\mathcal C &=& S + \sum_a l_a \left ( \sum_b p_aq_{ab} - p_a \right) + \sum_b m_b \left ( \sum_a p_aq_{ab} - p_b \right) \nonumber \\
&-& \frac{1}{2\varepsilon^2}\left (\sum_{a,b} p_a q_{ab} d(a,b)^2 -  \bar d^2  \right )  + \delta \left ( \sum_{a,b} p_a k_{ab} - 1\right )
\end{eqnarray}
Differentiating with respect to $q_{ab}$,
\begin{eqnarray}
0 &=& -(\log q_{ab} + 1) + l_a +  m_b -  \frac{d^2(a,b)}{2\varepsilon^2}\\
\Rightarrow q_{ab} &=& \frac{\rho_a \lambda_b}{p_a}\exp \left( -\frac{d(a,b)^2}{2\varepsilon^2}\right ) + \delta \label{eq:a}
\end{eqnarray}
Differentiating with respect to $p_a$
\begin{eqnarray}
0 &=& -\sum_{b} q_{ab} \log q_{ab} + l_a \sum_b q_{ab} - l_a + \sum_b m_b q_{ab} - m_a  \nonumber \\
&-& \frac{1}{2\varepsilon^2}\sum_b q_{ab} d(a,b)^2 + \delta \sum_b q_{ab} \label{eq:b}
\end{eqnarray}

From Eq.~\ref{eq:a} and Eq.~\ref{eq:b}, we have
\begin{eqnarray}
l_a + m_a = 1.
\end{eqnarray}
Thus,
\begin{eqnarray}
q_{ab} = \frac{\nu_{1b}}{\eta_1 \nu_{1a}} \exp \left( -\frac{d(a,b)^2}{2\varepsilon^2}\right )
\end{eqnarray}
where $\nu_{1a} = e^{-l_a}$ and $\eta = e^{-\delta}$. Imposing the normalization condition $\sum_b q_{ab} = 1$ identifies $\bar \nu_{1}$ as the Perron-Frobenius eigenvector of $\Delta$ and $\eta$ the corresponding eigenvalue~\citep{dixit2015stationary}.

\end{document}